\pdfoutput=1

\documentclass[11pt]{article}

\usepackage{naacl2021}
\usepackage{url}            
\usepackage{booktabs}       
\usepackage{amsfonts}       
\usepackage{nicefrac}       
\usepackage{microtype}      
\usepackage{subcaption}
\usepackage{amsmath,amsfonts,amssymb}
\usepackage{breqn}
\usepackage{tabularx}
\usepackage{multirow}
\usepackage{graphicx}
\usepackage{color}
\usepackage{tcolorbox}
\usepackage{CJK}
\usepackage{adjustbox}
\usepackage{xcolor}
\usepackage{colortbl}
\usepackage{multicol}
\usepackage{vwcol} 
\newsavebox{\algleft}
\newsavebox{\algright}
\usepackage{bm}
\usepackage{booktabs}
\usepackage{ctable} 
\usepackage{amsmath,stackengine}
\usepackage[linesnumbered,ruled,vlined]{algorithm2e}

\usepackage{times}
\usepackage{latexsym}

\usepackage[T1]{fontenc}

\usepackage[utf8]{inputenc}

\usepackage{microtype}

\title{Adapting BERT for Continual Learning of a Sequence of \\ Aspect Sentiment Classification Tasks}

\author{
Zixuan Ke$^{1}$, Hu Xu$^{2}$ \and Bing Liu$^{1}$ \\ $^1$Department of Computer Science, University of Illinois at Chicago\\
$^2$Facebook AI Research\\
$^1$\texttt{\{zke4,liub\}@uic.edu}\\  $^2$\texttt{huxu@fb.com}}

\begin{document}
\maketitle
\begin{abstract}

This paper studies continual learning (CL) of a sequence of aspect sentiment classification (ASC) tasks. Although some CL techniques have been proposed for document sentiment classification, we are not aware of any CL work on ASC. 
A CL system that incrementally learns a sequence of ASC tasks should address the following two issues:  
(1) transfer knowledge learned from previous tasks to the new task to help it learn a better model, and (2) maintain the performance of the models for previous tasks so that they are not forgotten. 
This paper proposes a novel capsule network based model called B-CL to address these issues.~B-CL markedly improves the ASC performance on both the new task and the old tasks via forward and backward knowledge transfer. The effectiveness of B-CL is demonstrated through extensive experiments.\footnote{https://github.com/ZixuanKe/PyContinual}
\end{abstract}

\section{Introduction}
\label{sec.intro}

Continual learning (CL) aims to incrementally learn a sequence of tasks. 
Once a task is learned, its training data is often discarded~\cite{chen2018lifelong}. This is in contrast to \textit{multi-task learning}, which assumes the training data of all tasks are available simultaneously. The CL setting is important in many practical scenarios. 
For example, a sentiment analysis company typically has many clients and each client often wants to have their private data deleted after use. In the personal assistant or chatbot context, the user does not want his/her chat data, which often contains sentiments or emotions, uploaded to a central server. 
In such applications, if we want to improve sentiment analysis accuracy for each user/client without breaching confidentiality, CL is a suitable solution. 

\begin{table}
    \centering
    \scalebox{0.567}{
        \begin{tabular}{l|l|l}
            \hline\hline
            Task ID & Domain/Task & One Training Example(in that domain/task)\\
            \hline\hline
            1 & Vacuum Cleaner [CF] & This vacuum cleaner \textit{sucks} !!! \\
            2 & Desktop [KT] & The keyboard is clicky .\\
            3 & Tablet [KT] & The soft keyboard is hard to use. \\
            \hline
            4 (new task) & Laptop & The new keyboard \textit{sucks} and is hard to click!\\
            \hline
        \end{tabular}
        \vspace{-1mm}
    }
    
    \caption{
    Tasks 2 and 3 have shareable \underline{k}nowledge to \underline{t}ransfer (KT) to the new task, whereas Task 1 has specific knowledge that is expected to be isolated from the new task to avoid \underline{c}atastrophic \underline{f}orgetting (CF) (although they use the same word).
    Note that here we use only one sentence to represent a task, but each task actually represents a domain with all its sentences.}
    \label{tbl:example}
    \vspace{-3mm}
\end{table}

There are two main types of continual learning: (1)~\textit{Task Incremental Learning} (TIL) and (2) \textit{Class Incremental Learning} (CIL). This work focuses on TIL, where each task is a separate \textit{aspect sentiment classification} (ASC) task. An ASC task is defined as follows~\cite{liu2015sentiment}: given an aspect (e.g., \textit{picture quality} in a camera review) and a sentence containing the aspect in a particular domain (e.g., camera), classify if the sentence expresses a positive, negative, or neutral (no opinion) about the aspect. 
{\color{black}TIL builds a model 
for each task and all models are in one neural network. 
In testing, the system knows which task each test instance belongs to and uses only the model for the task to classify the instance. In CIL, each task contains one or more classes to be learned. Only one model is built for all classes.} 
In testing, a test case from any class may be presented to the model to classify without giving it any task information. This setting is not applicable to ASC. 

Our goal of this paper is to achieve the following two objectives: (1) transfer the knowledge learned from previous tasks to the new task to help learn a better model for the new task without accessing the training data from previous tasks (in contrast to multi-task learning), and (2) maintain (or even improve) the performance of the old models for previous tasks so that they are not forgotten. 
The focus of the existing CL (TIL or CIL) research has been on solving (2),  \textit{catastrophic forgetting} (CF)~\cite{chen2018lifelong,ke2020ContinualMix}. CF means that when a network learns a sequence of tasks, the learning of each new task is likely to change the network parameters learned for previous tasks, which degrades the model performance for the previous tasks~\cite{mccloskey1989catastrophic}.
In our case, (1) is also important as ASC tasks are similar, i.e., words and phrases used to express sentiments for different products/tasks are similar.
To achieve the objectives, the system needs to identify the \textit{shared knowledge} that can be transferred to the new task to help it learn better and the \textit{task specific knowledge} that needs to be protected to avoid forgetting of previous models.  Table~\ref{tbl:example} gives an example.  
Fine-tuned BERT~\cite{DBLP:conf/naacl/DevlinCLT19} is one of the most effective methods for ASC~\cite{DBLP:conf/naacl/XuLSY19,sun-etal-2019-utilizing}. However, our experiments show that it works very poorly for TIL. The main reason is that the fine-tuned BERT on a task/domain captures highly task specific information which is difficult to transfer to a new task. 

In this paper, we propose a novel model called B-CL (\textit{BERT-based Continual Learning}) 
for ASC continual learning. The key novelty is a building block, called \textit{\underline{C}ontinual \underline{L}earning \underline{A}dapter} (CLA) inspired by the Adapter-BERT in~\cite{DBLP:conf/icml/HoulsbyGJMLGAG19}.~CLA leverages capsules and dynamic routing~\cite{sabour2017dynamic} to identify previous tasks that are similar to the new task and exploit their shared knowledge to help the new task learning and uses task masks to protect task-specific knowledge to avoid forgetting (CF). 
We conduct extensive experiments over a wide range of baselines to demonstrate the effectiveness of B-CL. 

In summary, this paper makes two key contributions. \textbf{(1)} 
    It proposes the problem of task incremental learning for ASC.
    \textbf{(2)} It proposes a new model B-CL with a novel adapter CLA incorporated in a pre-trained BERT to enable ASC continual learning. CLA employs capsules and dynamic routing to explore and transfer relevant knowledge from old tasks to the new task and uses task masks to isolate task-specific knowledge to avoid CF. To our knowledge, none of these has been done before. 

\section{Related Work}
\label{sec.related.work}

Continual learning (CL) has been studied extensively~\cite{chen2018lifelong,Parisi2019continual}. To our knowledge, no existing work has been done on CL for a sequence of ASC tasks, although CL of a sequence of document sentiment classification tasks has been done. 

\vspace{1mm}
\noindent
\textbf{Continual Learning.}
Existing work has mainly focused on dealing with catastrophic forgetting (CF). 

\textit{Regularization-based methods,} such as those in~\cite{Kirkpatrick2017overcoming,DBLP:conf/nips/LeeKJHZ17,Seff2017continual}, add a regularization in the loss to consolidate previous knowledge when learning a new task. 

\textit{Parameter isolation-based methods,} such as those in \cite{Serra2018overcoming,Mallya2017packnet,fernando2017pathnet}, make different subsets of the model parameters dedicated to different tasks and identify 
and mask them out during the training of the new task.

\textit{Gradient projection-based method}, such as that in~\cite{zeng2019continuous}, ensures the gradient updates occur only in the orthogonal direction to the input of the old tasks and thus will not affect old tasks. 

\textit{Replay-based methods}, such as those in~\cite{Rebuffi2017,Lopez2017gradient,Chaudhry2019ICLR}, retain an exemplar set of old task training data to help train the new task. The methods in~\cite{Shin2017continual,Kamra2017deep,Rostami2019ijcai,He2018overcoming} build data generators for previous tasks so that in learning the new task, they can use some generated data for previous tasks to help avoid forgetting. 

As these methods are mainly for avoiding CF, after learning a sequence of tasks, their final models are typically worse than learning each task separately. The proposed B-CL not only deals with CF, but also performs knowledge transfer to improve the performance of both the new and the old tasks.

\textbf{Lifelong Learning (LL).} LL is now regarded the same as CL, but early LL mainly aimed at improving the new task learning through forward transfer without tackling CF~\cite{Silver2013,ruvolo2013ella,chen2018lifelong}. 

Several researchers have used LL for document-level sentiment classification.  
\citet{DBLP:conf/acl/ChenM015} and~\citet{hao2019forward} proposed two Naive Bayes (NB) approaches to help improve the new task learning. A heuristic NB method was also used in~\cite{hao2019forward}.  
\citet{xia2017distantly} presented {\color{black}a} LL approach based on voting of individual task classifiers. 
All these works do not use neural networks, and are not concerned with the CF problem. 

\citet{ShuXuLiu2017} used LL for aspect extraction, which is a different problem.  \citet{shuai2018lifelong} used LL for ASC, but improved only the new task and did not deal with CF. 
Existing CL systems SRK~\cite{DBLP:conf/dasfaa/LvWLCZ19}, KAN \cite{ke2020continual} and L2PG~\cite{qin2020using} are for document sentiment classification, but not ASC. \citet{ke2020ContinualMix} also performed transfer in the image domain. 

Recently, capsule networks \cite{hinton2011transforming} have been used in sentiment classification and text classification \cite{DBLP:conf/acl/ChenQ19,DBLP:conf/acl/ZhaoPECY19}. But they have not been used in CL.


\section{Preliminary}
\label{sec.preliminary}
This section introduces BERT, Adapter-BERT and Capsule Network as they are used in our model. 

\textbf{BERT for ASC.} 
Due to its superior performance, this work uses BERT \cite{DBLP:conf/naacl/DevlinCLT19} and its transformer \cite{vaswani2017attention} architecture as the base.
We also adopt the ASC formulation in \cite{DBLP:conf/naacl/XuLSY19}, where the aspect term and review sentence are concatenated via \texttt{[SEP]}. The sentiment polarity is predicted on top of the \texttt{[CLS]} token.
Although BERT can achieve impressive performance on a single ASC task, its architecture and fine-tuning paradigm are not suitable for CL (see Sec.~\ref{sec.intro}). 
Experiments show that it performs very poorly for CL (Sec.~\ref{sec:results}).~We found that Adapter-BERT~\cite{DBLP:conf/icml/HoulsbyGJMLGAG19} is a better fit for CL.  

\begin{figure}[t]
\centering
\includegraphics[width=\columnwidth]{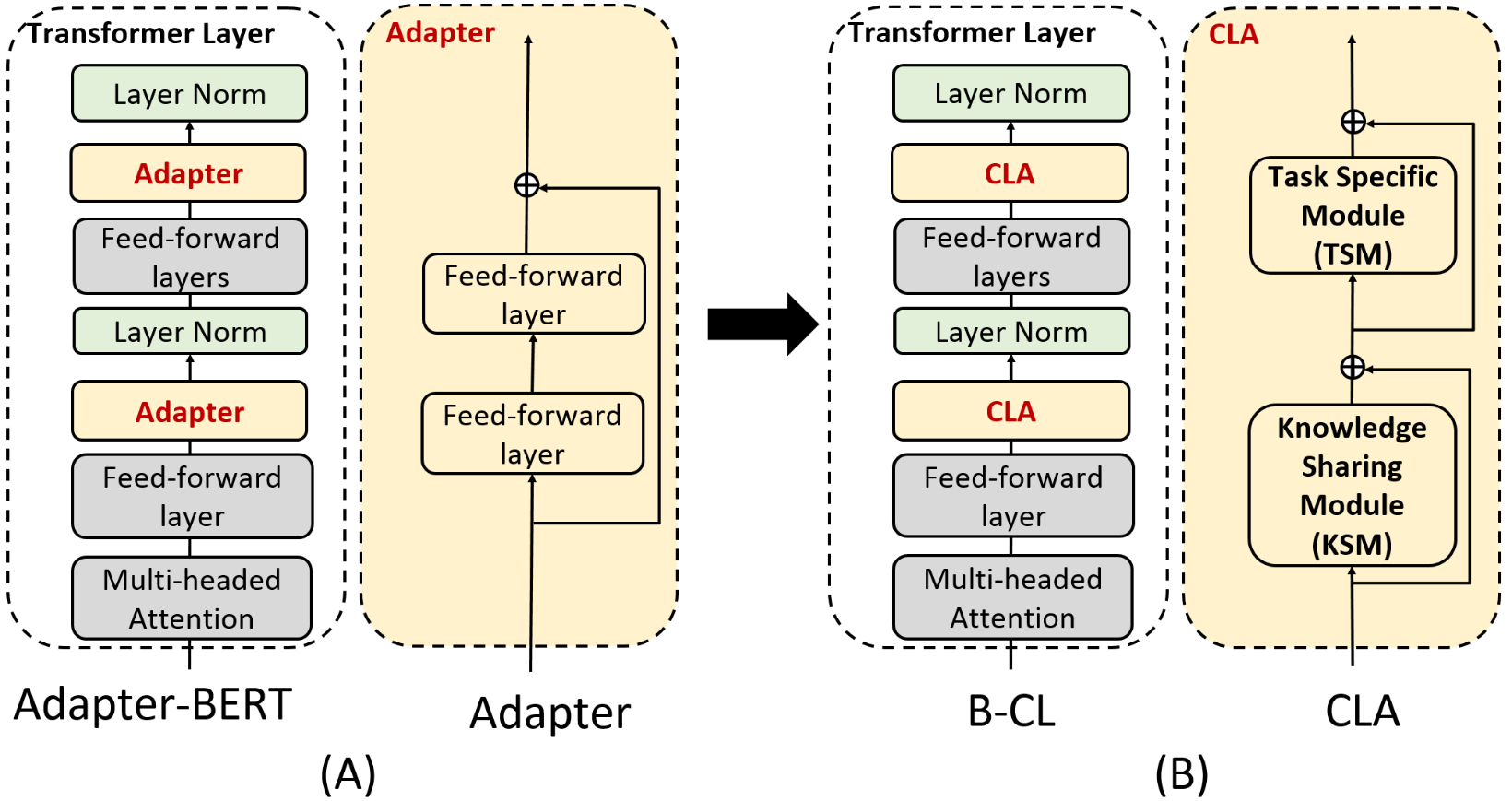}
\caption{
\textbf{(A).} Adapter-BERT \cite{DBLP:conf/icml/HoulsbyGJMLGAG19} and its adapters in a transformer~\cite{vaswani2017attention} layer. An adapter is a 2-layer fully connected network with a skip-connection. {\color{black}It is added twice to each Transformer layer. 
Only the adapters (yellow boxes) and layer norm (green boxes) layers are trainable. The other modules (grey boxes) are frozen.}
\textbf{(B).} Proposed B-CL, which replaces the adapter with CLA. CLA has two sub-modules: knowledge sharing module (KSM) and task specific module (TSM). Each of these modules has a skip-connection. 
}
\label{overview_adapter}
\vspace{-2mm}
\end{figure}

\textbf{Adapter-BERT.} 
Adapter-BERT basically inserts a 2-layer fully-connected network (adapter) in each transformer layer of BERT (see Figure~\ref{overview_adapter}(A)). During training for the end-task, only the adapters and normalization layers are trained, no change to any other BERT parameters, which is good for CL because fine-tuning BERT itself causes serious forgetting. Adapter-BERT achieves similar performances to fine-tuned BERT \cite{DBLP:conf/icml/HoulsbyGJMLGAG19}. We propose to exploit the adapter idea and the capsule network to achieve effective CL for ASC tasks. 

\textbf{Capsule Network.}
Capsule network (CapsNet) is a relatively new classification architecture  \cite{hinton2011transforming,sabour2017dynamic}. Unlike CNN, CapsNet replaces the scalar feature detectors with vector capsules that can preserve additional information such as position and thickness in images. A typical CapsNet has two capsule layers. The primary layer stores low-level feature maps and the class layer produces the probability for classification with each capsule corresponding to one class. 
{\color{black}It uses a dynamic routing algorithm to enable each lower level capsule to send its output to the similar (or ``agreed'', {\color{black}computed by dot product) } higher level capsule.} This is the key property that we exploit to identify and group similar tasks and their shared features or knowledge.


Note that the proposed B-CL does not adopt the whole capsule network as we are only interested in the capsule layers and dynamic routing instead of the max-margin loss and the classifier. 


\section{Continual Learning Adapter (CLA)}
\vspace{-1mm}

Recall the proposed B-CL aims to achieve (1) knowledge transfer between related old tasks and the new task through knowledge sharing and (2) forgetting avoidance through preventing task specific knowledge of previous tasks from being overwritten by the new task learning. 
Inspired by Adapter-BERT, we propose the \textit{continual learning adapters} (CLA) to replace the adapters in Adapter-BERT to enable CL as in Figure~\ref{overview_adapter}(B) to achieve BERT based continual learning for ASC. 

\begin{figure*}[t]
\centering
\includegraphics[width=\textwidth]{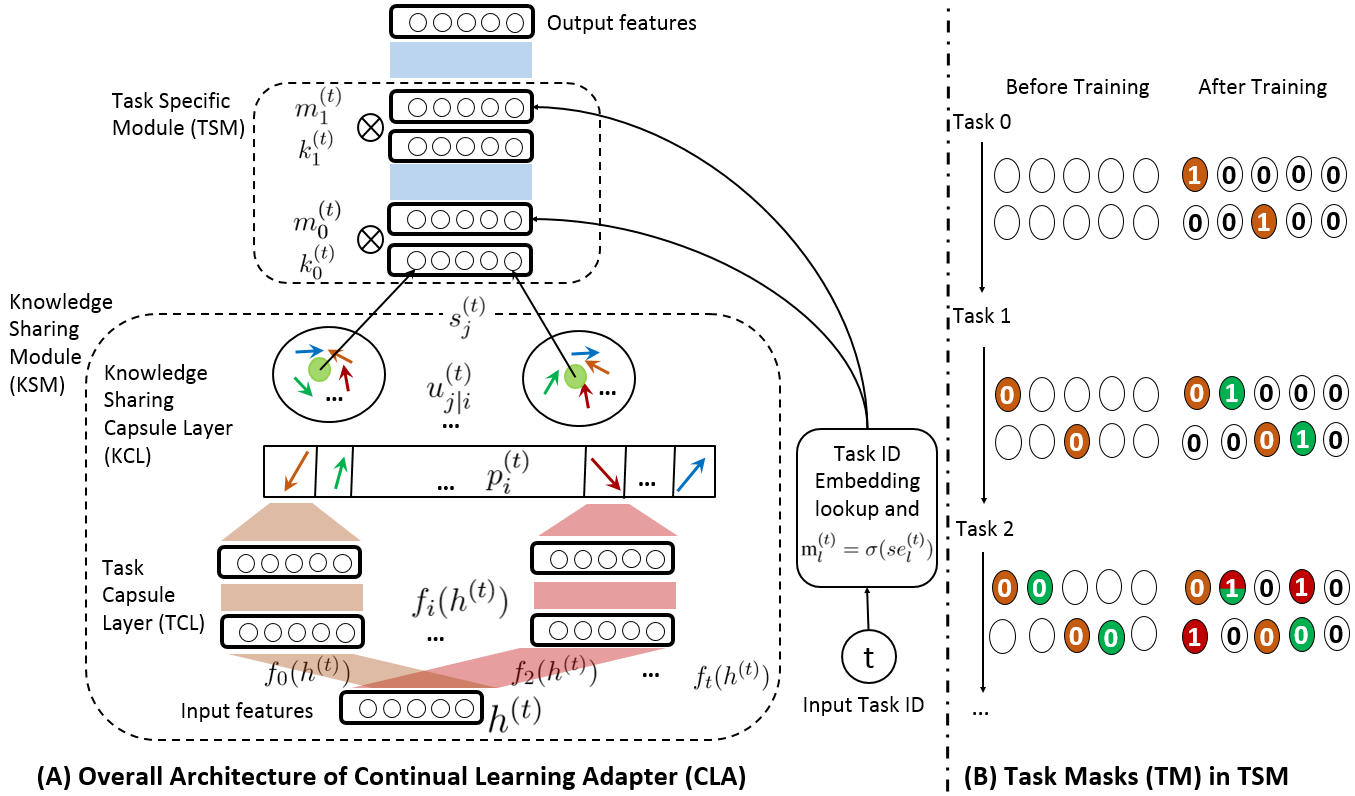}
\caption{
(A) Architecture of CLA: the skip-connection is not shown for clarity. (B) illustration of task masking: a (learnable) task mask is applied after the activation function to \textit{selectively} activate a neuron (or feature). Some
notes about (B) are: the two rows of each task corresponds to $k_0^{(t)}$ and $k_1^{(t)}$ in TSM. In the cells 
before training, those with 0’s are the neurons to be protected (masked) and those
cells without a number are free neurons (not used). In the cells after training, those cells with 1’s show
neurons that are important for the current task, which are used as a mask for the future. Those cells with more than one color indicate that they are shared by more than one task. 
Those 0 cells without a color are not used by any task.}
\vspace{-2mm}
\label{overview}
\end{figure*}

The architecture of CLA is shown in Figure~\ref{overview}(A). It contains two modules: (1) \textit{knowledge sharing module} (KSM) for identifying and exploiting shareable knowledge from the similar previous tasks and the new task, and (2) \textit{task specific module} (TSM) for learning task specific neurons and protecting them from being updated by the new task. 

CLA takes two inputs: 
(1) hidden states $h^{(t)}$ from the feed-forward layer inside a transformer layer and (2) task ID $t$.
The outputs are hidden states with features good for the $t$-th task. 
KSM leverages capsule layers (see below) and dynamic routing to group similar tasks and the shareable knowledge, whereas
TSM takes advantage of task mask (TM) to protect neurons for a particular task and leave other neurons free. Those free neurons are later used by TSM for a new task.
Since TMs are differentiable, the {\color{black}whole system B-CL can be trained end-to-end.}
We detail each module below.


\subsection{Knowledge Sharing Module (KSM)}
\label{sec:task_sharing}
\vspace{-0.5mm}



KSM groups similar tasks and shared knowledge (features) among them to enable knowledge transfer among similar tasks.~This is achieved through two capsule layers (\textit{task capsule layer} and \textit{knowledge sharing capsule layer}) and the dynamic routing algorithm of the capsule network.

\subsubsection{Task Capsule Layer (TCL)}
Each capsule in TCL represents a task and TCL prepares low-level features derived from each task (Figure~\ref{overview}(A)). As such, a capsule is added to TCL for every new task. This incremental growing is efficient and easy because these capsules are discrete and do not share parameters.
Also each capsule is simply a 2-layer fully connected network with a small number of parameters. 
Let $h^{(t)}\in\mathbb{R}^{d_t\times d_e}$ be the input of CLA, where $d_t$ is the number of tokens and $d_e$ the number of dimensions. Let the set of tasks learned so far be $\mathcal{T}_{\text{prev}}$ (before learning the new task $t$) and $|\mathcal{T}_{prev}|=n$. In TCL, we have $n+1$ different capsules representing all past $n$ learned tasks as well as the new task $t$. 
The capsule for the $i$-th ($i\leq n+1$) task is
\begin{equation}
\label{eq:task_specific}
p^{(t)}_i = f_i(h^{(t)}),
\end{equation}
where $f_i(\cdot)=\text{MLP}_i(\cdot)$ denotes a 2-layer fully-connected network.

\subsubsection{Knowledge Sharing Capsule Layer (KCL)}
Each \textit{knowledge sharing capsule} in KCL captures those tasks (i.e., their task capsules $\{p^{(t)}_i\}_1^{n+1}$) with similar features or shared knowledge.  
This is automatically achieved by the \textit{dynamic routing} algorithm. Recall dynamic routing encourages each lower level capsule (task capsule in our case) to send its output to the similar (or "agreed") higher level capsule (knowledge sharing capsule in our case).  

Essentially, the similar task capsules (with many shared features) are ``clustered'' together by higher coefficients (which determine how much a task capsule can go to the next layer) while dissimilar tasks (with few shared features) are blocked via low coefficients. Such clustering identifies the shared features or knowledge from multiple task capsules as well as helps backward transfer across the similar tasks.

KCL first turns each task capsule $p_{i}^{(t)}$ into a temporary feature $u_{j|i}^{(t)}$ 
as:
\begin{equation}
\label{eq:temp_task_shared_features}
u^{(t)}_{j|i} = W_{ij}p_{i}^{(t)},
\end{equation}
where $W_{ij}\in\mathbb{R}^{d_s\times d_k}$ is the weight matrix, $d_s$ and $d_k$ are the dimensions of task capsule $i$ and knowledge sharing capsule $j$. The number of knowledge sharing capsules is a hyperparameter detailed in the experiment section.
The temporary features are summed up with weights $c_{ij}^{(t)}$ to obtain the initial knowledge sharing capsule $s_j^{(t)}$:
\begin{equation}
\vspace{-1.5mm}
\label{eq:task_shared_features}
s^{(t)}_{j} = \sum_{i}c_{ij}^{(t)}u_{j|i}^{(t)},
\end{equation}
where $c_{ij}^{(t)}$ is a coupling coefficient summed up to 1 and we detail how to compute it later.
Note that the task capsule for each task in Eq.~\ref{eq:task_specific} is mapped to the knowledge sharing capsule in Eq.~\ref{eq:task_shared_features} and $c_{ij}^{(t)}$ indicates how much or how informative the representation of the $i$-th task is to the $j$-th knowledge sharing capsule.
As a result, a knowledge sharing capsule can represent diverse sharable knowledge.
For those tasks with a very low $c_{ij}^{(t)}$, their representations are less considered in the $j$-th knowledge sharing capsule. 
This makes sure only task capsules for tasks that are salient or similar to the new task are used and the others task capsules are ignored (and thus protected) to learn more general shareable knowledge.
Recall that the ASC tasks are similar and thus such learning of task sharing features can be very important. 

Note that in backpropagation, the dissimilar tasks with low $c_{ij}^{(t)}$ are updated with a low gradient while the similar tasks with high $c_{ij}^{(t)}$ are updated with a larger gradient. This encourages \textit{backward transfer} across similar tasks.

\textbf{Dynamic Routing.}
The coupling coefficient in Eq.~\ref{eq:task_shared_features} is essential for the quality of shareable knowledge.
This is computed by a ``routing softmax":
\vspace{-1.5mm}
\begin{equation}
\label{eq.routing_softmax}
c^{(t)}_{ij} = \frac{\exp(b^{(t)}_{ij})}{\sum_o\exp(b^{(t)}_{io})},
\end{equation}
where each $b_{ij}$ is the log prior probability showing how salient or similar a task capsule $i$ is to a knowledge sharing capsule $j$. It is initialized to 0 indicating no salient connection between them  
at the beginning.
We apply the dynamic routing algorithm in~\cite{sabour2017dynamic} to update $b_{ij}$:
\begin{equation}
\label{eq.coupling_coefficients}
b^{(t)}_{ij} \leftarrow b_{ij}^{(t)} + a_{ij}^{(t)},
\end{equation}
where $a_{ij}$ is the agreement coefficient (see below).
Intuitively, this step tends to aggregate the similar (or ``agreed'') tasks on a knowledge sharing capsule 
with a higher agreement coefficient $a_{ij}$ 
and thus a higher logit $b_{ij}^{(t)}$ (Eq.~\ref{eq.coupling_coefficients})
or coupling coefficient $c_{ij}^{(t)}$ (Eq.~\ref{eq.routing_softmax}).
The agreement coefficient is computed as 
\begin{equation}
\vspace{-1.5mm}
\label{eq.agreement_coefficient}
a^{(t)}_{ij} = u_{j|i}^{(t)} \cdot v_j^{(t)},
\end{equation}
where $v^{(t)}_j$ is a normalized representation by applying the non-linear ``squash" function \cite{sabour2017dynamic} to $s^{(t)}_j$ (for the first task, $s_j^{(t)}=u_{j|i}^{(t)}$):
\vspace{-1.5mm}
\begin{equation}
\vspace{-1.5mm}
\label{eq.squash}
v^{(t)}_j = \frac{||s_j^{(t)}||^2}{1+||s_j^{(t)}||}\frac{s_j^{(t)}}{||s_j^{(t)}||},
\end{equation}
where the length of $v^{(t)}_j$ is normalized to [0,1] to represent the active probability of a knowledge sharing capsule $j$. 

Finally, note that the dynamic routing procedure 
(Eq.~(\ref{eq:task_shared_features})$\to$(\ref{eq.squash}))
is repeated for $r$ iterations.

\subsection{Task Specific Module (TSM)} 
Although knowledge sharing is important for ASC, it is equally important to preserve task specific knowledge for previous tasks to prevent forgetting (CF). 
To achieve this, we use task masks (Figure~\ref{overview}(B)). Specifically, we first detect the neurons used by each old task, and then block off or mask out all the \textit{used} neurons when learning a new task. 

The task specific module consists of 
differentiable layers (CLA uses a 2-layer fully-connected network).
Each layer's output is further applied with a task mask to indicate which neurons should be protected for that task to overcome CF and forbids gradient updates for those neurons during backpropagation for a new task. 
Those tasks with overlapping masks indicate knowledge sharing. 
Due to KSM, the features flowing in those overlapping neurons enable the related old tasks to also improve in learning the new task.  

\subsection{Task Masks}
Given the knowledge sharing capsule $s_j^{(t)}$, TSM maps them into input $k_l^{(t)}$ via a fully-connected network, where $l$ is the $l$-th layer in TSM. 
A task mask (a ``soft'' binary mask) $\text{m}^{(t)}_l$ is trained for each task $t$ at each layer $l$ in TSM during training task $t$'s classifier, indicating the neurons that are important for the task in the layer. Here we borrow the hard attention idea in \cite{Serra2018overcoming} and leverage the task ID embedding to the train the task mask.

For a task ID $t$, its embedding $e^{(t)}_l$ consists of differentiable deterministic parameters that can be learned together with other parts of the network. 
It is trained for each layer in TSM.
To generate the task mask $\text{m}^{(t)}_l$ from $e^{(t)}_l$, \textit{Sigmoid} is used as a pseudo-gate function and a positive 
scaling hyper-parameter $s$ is applied to help training. The $m^{(t)}_l$ is computed as follows:
\vspace{-1.5mm}
\begin{equation}
\vspace{-1.5mm}
\label{eq1}
m^{(t)}_l = \sigma(se^{(t)}_l).
\end{equation}
Note that the neurons in $m^{(t)}_l$ may overlap with those in other $m^{(i_{\text{prev}})}_l$s from previous tasks showing some shared knowledge. Given the output of each layer in TSM, $k_l^{(t)}$, we element-wise multiply $k_l^{(t)} \otimes m^{(t)}_l$. The masked output of the last layer $k^{(t)}$ is fed to the next layer of the BERT with a skip-connection (see Figure~\ref{overview_adapter}). After learning task $t$, the final $m^{(t)}_l$ is saved and added to the set $\{m^{(t)}_l\}$. 

\subsection{Training} 
For each past task $i_{\text{prev}} \in \mathcal{T}_{\text{prev}}$, its mask $m^{(i_{\text{prev}})}_l$ 
indicates which neurons are used by that task and need to be protected. 
In learning task $t$, $m^{(i_{\text{prev}})}_l$ is used to set the gradient $g^{(t)}_l$ on \textit{all} used neurons of the layer $l$ in TSM to 0. Before modifying the gradient, we first accumulate all used neurons by all previous tasks' masks.
Since $m^{(i_{\text{prev}})}_l$ is binary, we use 
max-pooling 
to achieve the accumulation:  
\vspace{-1.5mm}
\begin{equation}
\vspace{-1.5mm}
m^{(t_{\text{ac}})}_l = \text{MaxPool}(\{m^{(i_{\text{prev}})}_l\}).
\end{equation}


The term $m^{(t_{\text{ac}})}_l$ is applied to the gradient:
\vspace{-1.5mm}
\begin{equation}
\vspace{-1.5mm}
g^{'(t)}_l = g^{(t)}_l \otimes (1-m^{(t_{\text{ac}})}_l).
\end{equation}
Those gradients corresponding to the 1 entries in $m^{(t_{\text{ac}})}_l$ are set to 0 while the others remain unchanged. 
In this way, neurons in an old task are protected. 
Note that we expand (copy) the vector $m_l^{(t_{\text{ac}})}$ to match the dimensions of $g_l^{(t)}$.

Though the idea is intuitive, $e^{(t)}_l$ is not easy to train. To make the learning of $e^{(t)}_l$ easier and more stable, an annealing strategy is applied~\citep{Serra2018overcoming}. That is, $s$ is annealed during training, inducing a gradient flow and set $s=s_{\max}$ during testing. {\color{black}Eq.~\ref{eq1} approximates a unit step function as the mask, with $m_l^{(t)} \to \{0, 1\}$ when $s \to \infty$. 
A training epoch starts with all neurons being equally active, 
which are progressively polarized within the epoch. Specifically, $s$ is annealed as follows}:
\vspace{-1.5mm}
\begin{equation}
\vspace{-1.5mm}
\label{eq:smax}
s = \frac{1}{s_{\max}} + (s_{\max} - \frac{1}{s_{\max}})\frac{b-1}{B-1},
\end{equation}
where $b$ is the batch index and $B$ is the total number of batches in an epoch.

\textbf{Illustration.} In Figure~\ref{overview}(B), after learning the first task (Task 0), we obtain its useful neurons marked in orange with a 1 in each neuron, which serves as a mask in learning future tasks. In learning task 1, 
those useful neurons for task 0 are masked (with 0 in those orange neurons or cells on the left). The process also learns the useful neurons for task 1 marked in green with 1's. When task 2 arrives, all important neurons for tasks 0 and 1 are masked, i.e., its mask entries are set to 0 (orange and green before training). After training task 2, we see that task 2 and task 1 have a shared neuron that is important to both of them. The shared neuron is marked in both red and green.

\section{Experiments}
We now evaluate B-CL by comparing it with both \textit{non-continual learning} and \textit{continual learning} baselines. We follow the standard CL evaluation method in~\citep{DBLP:journals/corr/abs-1909-08383}.
We first present B-CL a sequence of aspect sentiment classification (ASC) tasks for it to learn. Once a task is learned, its training data is discarded. After all tasks are learned, we test all task models using their respective test data. 
In training each task, we use its validation set to decide when to stop training. 

\subsection{Experiment Datasets}

\begin{table}[]
\centering
\resizebox{0.9\columnwidth}{!}{
\begin{tabular}{ccccc}
\specialrule{.2em}{.1em}{.1em}
Data source & Task/domain & Train & Validation & Test \\
\specialrule{.1em}{.05em}{.05em}

\multirow{3}{*}{Liu3domain} & Speaker & 352  & 44 & 44 \\
 & Router & 245 & 31 & 31 \\
 & Computer & 283 & 35 & 36  \\
\specialrule{.1em}{.05em}{.05em}
\multirow{5}{*}{HL5domain} & Nokia6610 & 271 & 34 & 34  \\
 & Nikon4300 & 162 & 20  & 21 \\
 & Creative & 677 & 85 & 85 \\
 & CanonG3 & 228 & 29 & 29 \\
 & ApexAD & 343 & 43 & 43 \\
\specialrule{.1em}{.05em}{.05em}
\multirow{9}{*}{Ding9domain} & CanonD500 & 118 & 15 & 15 \\
 & Canon100 & 175 & 22 & 22 \\
 & Diaper & 191 & 24 & 24 \\
 & Hitachi & 212 & 26 & 27\\
 & Ipod & 153 & 19 & 20 \\
 & Linksys & 176 & 22 & 23 \\
 & MicroMP3 & 484 & 61 & 61 \\
 & Nokia6600 & 362 & 45 & 46 \\
 & Norton & 194 & 24 & 25 \\
\specialrule{.1em}{.05em}{.05em}
\multirow{2}{*}{SemEval14} & Rest. & 3452 & 150 & 1120 \\
 & Laptop & 2163 & 150 & 638 \\
\specialrule{.1em}{.05em}{.05em}

\end{tabular}
}
\caption{Number of examples in each task or dataset. More detailed data statistics are given in the \textit{Appendix}.}
\label{tab:dataset_main}
\vspace{-3mm}
\end{table}
{\color{black}Since B-CL works in the 
CL setting, we employ a set of 19 ASC datasets (reviews of 19 products) to produce sequences of tasks. Each dataset represents a task. The datasets are from 4 sources:} (1) \textbf{HL5Domains} \cite{hu2004mining} with reviews of 5 products; (2) \textbf{Liu3Domains} \cite{liu2015automated} with reviews of 3 products; (3) \textbf{Ding9Domains} \cite{ding2008holistic} with reviews of 9 products; and (4) \textbf{SemEval14} with reviews of 2 products - SemEval 2014 Task 4 for laptop and restaurant. 
For (1), (2) and (3), we split about 10\% of the original data as the validation data, another about 10\% of the original data as the testing data. For (4), we use 150 examples from the training set for validation. To be consistent with existing research \cite{tang-etal-2016-aspect},
examples belonging to the conflict polarity (both positive and negative sentiments are expressed about an aspect term) are not used.
Statistics of the 19 datasets are given in Table \ref{tab:dataset_main}.

\subsection{Compared Baselines}
\label{sec:baselines}

We use 18 baselines, including both \textit{non-continual learning} and \textit{continual learning} methods. 

\textbf{Non-continual Learning (NL) Baselines}: NL setting builds a model for each task independently using a separate network. It clearly has no knowledge transfer or forgetting. We have 3 baselines under NL, \textbf{(1) BERT}, \textbf{(2) Adapter-BERT} and \textbf{(3) W2V} (word2vec embeddings). For \textbf{BERT}, we use trainable BERT to perform ASC (see Sec.~\ref{sec.preliminary}); \textbf{Adapter-BERT} adapts the BERT as in \cite{DBLP:conf/icml/HoulsbyGJMLGAG19}, where only the adapter blocks 
are trainable; 
\textbf{W2V} uses embeddings 
trained on the Amazon review data in \cite{Xu2018pro} using FastText \cite{grave2018learning}. We adopt the ASC classification network in~\cite{DBLP:conf/acl/LiX18}, which takes both aspect term and review sentence as input. 

\textbf{Continual Learning (CL) Baselines}. {\color{black}CL setting includes 3 baselines \textit{without dealing with forgetting} (\textbf{WDF}) and 12 baselines from 6 state-of-the art \textit{task incremental learning} (TIL) methods dealing with forgetting. WDF baselines 
greedily learn a sequence of tasks incrementally without explicitly tackling forgetting or knowledge transfer. The 3 baselines under WDF are also \textbf{(4) BERT}, \textbf{(5) Adapter-BERT} and \textbf{(6) W2V}.

The 6 state-of-the-art CL systems are:} KAN, SRK, HAT, UCL, EWC and OWM. 
\textbf{KAN} \citep{ke2020continual} and \textbf{SRK} \citep{DBLP:conf/dasfaa/LvWLCZ19} are  TIL methods 
for document sentiment classification. 
HAT, UCL, EWC and OWM were originally designed for image classification. We replace their original MLP or CNN image classification network with CNN for text classification \cite{DBLP:conf/emnlp/Kim14}. 
\textbf{HAT} 
\citep{Serra2018overcoming} is one of the best TIL methods with almost no forgetting. 
\textbf{UCL}
\citep{DBLP:conf/nips/AhnCLM19} is a latest TIL method. 
\textbf{EWC} \citep{Kirkpatrick2017overcoming} is a popular regularization-based class incremental learning (CIL) method, which was adapted 
for TIL by only training on the corresponding head of the specific task ID during training and only considering the corresponding head's prediction during testing.
\textbf{OWM} \citep{zeng2019continuous} is a state-of-the-art CIL method, which we also adapt to TIL. 

From the 6 systems, we created \textbf{6 baselines} using \textbf{W2V} embeddings with the aspect term added before the sentence so that the CL methods can take both aspect and the review sentence, and \textbf{6 baselines} using \textbf{BERT (Frozen)} (which replaces W2V embeddings). Following the BERT formulation in Sec.~\ref{sec.preliminary}, it can naturally take both aspect and review sentence. 
Adapter-BERT is not applicable to them as their architecture cannot use an adapter.

\subsection{Hyperparameters}
Unless otherwise stated, for the task sharing module, we employ 2 layers of fully connected network with dimensions 768 in TCL. We also employ 3 knowledge sharing capsules. The dynamic routing is repeated for 3 iterations. For the task-specific module, We employ the embedding with 2000 dimensions as the final and hidden layer of the TSM. The task ID embeddings have 2000 dimensions. A fully connected layer with softmax output is used as the classification heads in the last layer of the BERT, together with the categorical cross-entropy loss. We use 140 for $s_{\max}$ in Eq.~\ref{eq:smax},
dropout of 0.5 between fully connected layers. The training of BERT, Adapter-BERT and B-CL follow that of \cite{DBLP:conf/naacl/XuLSY19}. We adopt $\text{BERT}_{\textbf{BASE}}$ (uncased). The maximum length of the sum of sentence and aspect is set to 128. We use Adam optimizer and set the learning rate to 3e-5.
For the SemEval datasets, 10 epochs are used and for all other datasets, 30 epochs are used based on results from validation data.
All runs use the batch size 32. For the CL baselines, we train all models with the learning rate of 0.05. 
We early-stop training when there is no improvement in the validation loss for 5 epochs. The batch size is set to 64. {\color{black}For all the CL baselines, we use the code provided by their authors and adopt their original parameters (for EWC, we adopt its TIL variant
implemented by \cite{Serra2018overcoming}).}

\begin{table}[]
\centering
\resizebox{\columnwidth}{!}{
\begin{tabular}{ccc||cc}
\specialrule{.2em}{.1em}{.1em}
Scenario & Category & Model & Acc. & MF1 \\
\specialrule{.1em}{.05em}{.05em}
\multirow{3}{*}{\begin{tabular}[c]{@{}c@{}}Non-continual\\      Learning\end{tabular}} & BERT & NL & 0.8584 & 0.7635 \\
 & Adapter-BERT & NL & 0.8596 & 0.7807 \\
 & W2V & NL & 0.7701 & 0.5189 \\
 \cline{1-5}
\multirow{18}{*}{\begin{tabular}[c]{@{}c@{}}Continual\\      Learning\end{tabular}} & BERT & WDF & 0.4960 & 0.4308 \\
 & Adapter-BERT & WDF & 0.5403 & 0.4481 \\
 & W2V & WDF & 0.8269 & 0.7356 \\
 \cline{2-5}
 & \multirow{6}{*}{\begin{tabular}[c]{@{}c@{}}BERT\\      (Frozen)\end{tabular}} & KAN & 0.8549 & 0.7738 \\
 &  & SRK & 0.8476 & 0.7852 \\
 &  & EWC & 0.8637 & 0.7452 \\
 &  & UCL & 0.8389 & 0.7482 \\
 &  & OWM & 0.8702 & 0.7931 \\
 &  & HAT & 0.8674 & 0.7816 \\
 \cline{2-5}
  & \multirow{6}{*}{W2V} & KAN & 0.7206 & 0.4001 \\		
 &  & SRK & 0.7101 & 0.3963 \\
 &  & EWC & 0.8416 & 0.7229 \\
 &  & UCL & 0.8441 & 0.7599 \\
 &  & OWM & 0.8270 & 0.7118 \\
 &  & HAT & 0.8083 & 0.6363 \\
  \cline{2-5}
  & \multicolumn{2}{c||}{\textbf{B-CL (forward)}} & \textbf{0.8809} & \textbf{0.7993} \\
 & \multicolumn{2}{c||}{\textbf{B-CL}} & \textbf{0.8829} & \textbf{0.8140} \\
\specialrule{.1em}{.05em}{.05em}
\end{tabular}
}
\caption{Accuracy (Acc.) and Macro-F1 (MF1) averaged over 5 random sequences of 19 tasks. 
\vspace{-2mm}
} 
\label{tab:overall_results}
\end{table}

\subsection{Results and Analysis}
\label{sec:results}
Since the order of the 19 tasks may have an impact on the final results, we randomly choose and run 5 task sequences and average their results. 
We compute both accuracy and Macro-F1 over 3 classes of polarities, 
where Macro-F1 is the major metric as the imbalanced classes introduce biases on accuracy. {\color{black}Table~\ref{tab:overall_results} gives the average results of 19 tasks (or datasets) over the 5 random task sequences.} 

\textbf{Overall Performance.} 
Table~\ref{tab:overall_results} shows that B-CL outperforms all baselines markedly. We discuss the detailed observations below:

(1)~For non-continual learning (NL) baselines, BERT and Adapter-BERT perform similarly. 
W2V is poorer, which is understandable.  

{\color{black}(2) Comparing NL (non-continual learning) and WDF (continual learning without dealing with forgetting), we see WDF is much better than NL for W2V. This indicates ASC tasks are similar and have shared knowledge. Catastrophic forgetting (CF) is not a major issue for W2V. 

However, WDF is much worse than NL for BERT (with fine-tuning) and Adapter-BERT (with adapter-tuning). This is because BERT with fine-tuning learns highly task specific knowledge \cite{DBLP:journals/corr/abs-2004-14448}. While this is desirable for NL, it is bad for WDF because task specific knowledge is hard to share across tasks or transfer. Then WDF 
causes serious forgetting (CF) for CL.} 

(3) Unlike BERT and Adapter-BERT, our B-CL can do very well in both forgetting avoidance and knowledge transfer (outperforming all baselines). 
For state-of-the-art CL baselines, EWC, UCL, OWM and HAT, although they perform better than WDF, they are all significantly poorer than B-CL as they don't have methods to encourage knowledge transfer. KAN and SRK do knowledge transfer but they are for document-level sentiment classification. They are weak, even weaker than other CL methods. 



\textbf{Effectiveness of Knowledge Transfer.}
We now look at knowledge transfer of B-CL. For forward transfer (B-CL(forward)) in Table~\ref{tab:overall_results}), we use the test accuracy and MF1 of each task when it was first learned. For backward transfer (B-CL in Table~\ref{tab:overall_results}), we use the final result after all tasks are learned. By comparing the results of NL with the results of forward transfer, we can see whether forward transfer is effective. By comparing the forward transfer result with the backward transfer result, we can see whether the backward transfer can improve further. 
{\color{black}The average results of B-CL forward (\textbf{B-CL(forward)}) and backward (\textbf{B-CL}) are given in Table \ref{tab:overall_results}. It shows that forward transfer of B-CL is highly effective} (forward results for other CL baselines are given in the \textit{Appendix} and we see B-CL's forward result outperforms all baselines' forward results). For backward transfer, B-CL slightly improves the performance.
\textbf{Ablation Experiments.} 
The results of ablation experiments are in Table \ref{tab:ablation_results}. ``-KSM;-TSM'' means without knowledge sharing and task specific modules, simply deploying an Adapter-BERT. ``-KSM'' means without the knowledge sharing module. ``-TSM'' means without the task specific module. Table \ref{tab:ablation_results} clearly shows that the full B-CL system always gives the best overall results, indicating every component contributes to the model. 



\begin{table}[]
\centering
\resizebox{0.7\columnwidth}{!}{
\begin{tabular}{c||cc}
\specialrule{.2em}{.1em}{.1em}
Model & Acc. & MF1  \\
\specialrule{.1em}{.05em}{.05em}
B-CL (-KSM;-TSM) & 0.5403 & 0.4481 \\
B-CL (-KSM)& 0.8614 & 0.7852  \\
B-CL (-TSM) & 0.8312 & 0.7107 \\
\textbf{B-CL} & \textbf{0.8829} & \textbf{0.8140} \\
\specialrule{.1em}{.05em}{.05em}
\end{tabular}
}
\caption{Ablation experiment results.}
\label{tab:ablation_results}
\vspace{-4mm}
\end{table}


\section{Conclusion}
This paper studies continual learning (CL) of a sequence of ASC tasks. 
It proposed a novel technique called B-CL that can be applied to pre-trained BERT for CL. B-CL uses continual learning adapters and capsule networks to effectively encourage knowledge transfer among tasks and also to protect task-specific knowledge.
Experiments show that B-CL markedly improves the ASC performance on both the new task and the old tasks via forward and backward knowledge transfer.

\section*{Acknowledgments}
This work was supported in part by two grants from National Science Foundation: IIS-1910424 and IIS-1838770, a DARPA Contract HR001120C0023, and a research gift from Northrop Grumman.





\bibliography{anthology,custom}

\begin{thebibliography}{49}
\expandafter\ifx\csname natexlab\endcsname\relax\def\natexlab#1{#1}\fi

\bibitem[{Ahn et~al.(2019)Ahn, Cha, Lee, and Moon}]{DBLP:conf/nips/AhnCLM19}
Hongjoon Ahn, Sungmin Cha, Donggyu Lee, and Taesup Moon. 2019.
\newblock \href
  {http://papers.nips.cc/paper/8690-uncertainty-based-continual-learning-with-adaptive-regularization}
  {Uncertainty-based continual learning with adaptive regularization}.
\newblock In \emph{NIPS}, pages 4394--4404.

\bibitem[{Chaudhry et~al.(2019)Chaudhry, Ranzato, Rohrbach, and
  Elhoseiny}]{Chaudhry2019ICLR}
Arslan Chaudhry, Marc'Aurelio Ranzato, Marcus Rohrbach, and Mohamed Elhoseiny.
  2019.
\newblock \href {https://openreview.net/forum?id=Hkf2\_sC5FX} {Efficient
  lifelong learning with {A-GEM}}.
\newblock In \emph{ICLR}.

\bibitem[{Chen and Liu(2018)}]{chen2018lifelong}
Zhiyuan Chen and Bing Liu. 2018.
\newblock Lifelong machine learning.
\newblock \emph{Synthesis Lectures on Artificial Intelligence and Machine
  Learning}, 12(3):1--207.

\bibitem[{Chen et~al.(2015)Chen, Ma, and Liu}]{DBLP:conf/acl/ChenM015}
Zhiyuan Chen, Nianzu Ma, and Bing Liu. 2015.
\newblock \href {https://www.aclweb.org/anthology/P15-2123/} {Lifelong learning
  for sentiment classification}.
\newblock In \emph{ACL}, pages 750--756.

\bibitem[{Chen and Qian(2019)}]{DBLP:conf/acl/ChenQ19}
Zhuang Chen and Tieyun Qian. 2019.
\newblock \href {https://doi.org/10.18653/v1/p19-1052} {Transfer capsule
  network for aspect level sentiment classification}.
\newblock In \emph{ACL}, pages 547--556. Association for Computational
  Linguistics.

\bibitem[{Devlin et~al.(2019)Devlin, Chang, Lee, and
  Toutanova}]{DBLP:conf/naacl/DevlinCLT19}
Jacob Devlin, Ming{-}Wei Chang, Kenton Lee, and Kristina Toutanova. 2019.
\newblock \href {https://doi.org/10.18653/v1/n19-1423} {{BERT:} pre-training of
  deep bidirectional transformers for language understanding}.
\newblock In \emph{NAACL-HLT}, pages 4171--4186. Association for Computational
  Linguistics.

\bibitem[{Ding et~al.(2008)Ding, Liu, and Yu}]{ding2008holistic}
Xiaowen Ding, Bing Liu, and Philip~S Yu. 2008.
\newblock A holistic lexicon-based approach to opinion mining.
\newblock In \emph{Proceedings of the 2008 international conference on web
  search and data mining}, pages 231--240.

\bibitem[{Fernando et~al.(2017)Fernando, Banarse, Blundell, Zwols, Ha, Rusu,
  Pritzel, and Wierstra}]{fernando2017pathnet}
Chrisantha Fernando, Dylan Banarse, Charles Blundell, Yori Zwols, David Ha,
  Andrei~A. Rusu, Alexander Pritzel, and Daan Wierstra. 2017.
\newblock \href {http://arxiv.org/abs/1701.08734} {Pathnet: Evolution channels
  gradient descent in super neural networks}.
\newblock \emph{CoRR}, abs/1701.08734.

\bibitem[{Grave et~al.(2018)Grave, Bojanowski, Gupta, Joulin, and
  Mikolov}]{grave2018learning}
Edouard Grave, Piotr Bojanowski, Prakhar Gupta, Armand Joulin, and Tomas
  Mikolov. 2018.
\newblock Learning word vectors for 157 languages.
\newblock In \emph{LREC}.

\bibitem[{He and Jaeger(2018)}]{He2018overcoming}
Xu~He and Herbert Jaeger. 2018.
\newblock \href {https://openreview.net/forum?id=B1al7jg0b} {Overcoming
  catastrophic interference using conceptor-aided backpropagation}.
\newblock In \emph{ICLR}.

\bibitem[{Hinton et~al.(2011)Hinton, Krizhevsky, and
  Wang}]{hinton2011transforming}
Geoffrey~E Hinton, Alex Krizhevsky, and Sida~D Wang. 2011.
\newblock Transforming auto-encoders.
\newblock In \emph{International conference on artificial neural networks},
  pages 44--51. Springer.

\bibitem[{Houlsby et~al.(2019)Houlsby, Giurgiu, Jastrzebski, Morrone,
  de~Laroussilhe, Gesmundo, Attariyan, and
  Gelly}]{DBLP:conf/icml/HoulsbyGJMLGAG19}
Neil Houlsby, Andrei Giurgiu, Stanislaw Jastrzebski, Bruna Morrone, Quentin
  de~Laroussilhe, Andrea Gesmundo, Mona Attariyan, and Sylvain Gelly. 2019.
\newblock \href {http://proceedings.mlr.press/v97/houlsby19a.html}
  {Parameter-efficient transfer learning for {NLP}}.
\newblock In \emph{ICML}, volume~97 of \emph{Proceedings of Machine Learning
  Research}, pages 2790--2799. {PMLR}.

\bibitem[{Hu and Liu(2004)}]{hu2004mining}
Minqing Hu and Bing Liu. 2004.
\newblock Mining and summarizing customer reviews.
\newblock In \emph{Proceedings of ACM SIGKDD}, pages 168--177.

\bibitem[{Kamra et~al.(2017)Kamra, Gupta, and Liu}]{Kamra2017deep}
Nitin Kamra, Umang Gupta, and Yan Liu. 2017.
\newblock \href {http://arxiv.org/abs/1710.10368} {Deep generative dual memory
  network for continual learning}.
\newblock \emph{CoRR}, abs/1710.10368.

\bibitem[{Ke et~al.(2020{\natexlab{a}})Ke, Liu, and Huang}]{ke2020ContinualMix}
Zixuan Ke, Bing Liu, and Xingchang Huang. 2020{\natexlab{a}}.
\newblock Continual learning of a mixed sequence of similar and dissimilar
  tasks.
\newblock In \emph{NeurIPS}.

\bibitem[{Ke et~al.(2020{\natexlab{b}})Ke, Liu, Wang, and
  Shu}]{ke2020continual}
Zixuan Ke, Bing Liu, Hao Wang, and Lei Shu. 2020{\natexlab{b}}.
\newblock Continual learning with knowledge transfer for sentiment
  classification.
\newblock In \emph{ECML-PKDD}.

\bibitem[{Kim(2014)}]{DBLP:conf/emnlp/Kim14}
Yoon Kim. 2014.
\newblock \href {https://doi.org/10.3115/v1/d14-1181} {Convolutional neural
  networks for sentence classification}.
\newblock In \emph{EMNLP}, pages 1746--1751. {ACL}.

\bibitem[{Kirkpatrick et~al.(2016)Kirkpatrick, Pascanu, Rabinowitz, Veness,
  Desjardins, Rusu, Milan, Quan, Ramalho, Grabska{-}Barwinska, Hassabis,
  Clopath, Kumaran, and Hadsell}]{Kirkpatrick2017overcoming}
James Kirkpatrick, Razvan Pascanu, Neil~C. Rabinowitz, Joel Veness, Guillaume
  Desjardins, Andrei~A. Rusu, Kieran Milan, John Quan, Tiago Ramalho, Agnieszka
  Grabska{-}Barwinska, Demis Hassabis, Claudia Clopath, Dharshan Kumaran, and
  Raia Hadsell. 2016.
\newblock \href {http://arxiv.org/abs/1612.00796} {Overcoming catastrophic
  forgetting in neural networks}.
\newblock \emph{CoRR}, abs/1612.00796.

\bibitem[{Lange et~al.(2019)Lange, Aljundi, Masana, and
  Tuytelaars}]{DBLP:journals/corr/abs-1909-08383}
Matthias~De Lange, Rahaf Aljundi, Marc Masana, and Tinne Tuytelaars. 2019.
\newblock \href {http://arxiv.org/abs/1909.08383} {Continual learning: {A}
  comparative study on how to defy forgetting in classification tasks}.
\newblock \emph{CoRR}, abs/1909.08383.

\bibitem[{Lee et~al.(2017)Lee, Kim, Jun, Ha, and
  Zhang}]{DBLP:conf/nips/LeeKJHZ17}
Sang{-}Woo Lee, Jin{-}Hwa Kim, Jaehyun Jun, Jung{-}Woo Ha, and Byoung{-}Tak
  Zhang. 2017.
\newblock \href
  {http://papers.nips.cc/paper/7051-overcoming-catastrophic-forgetting-by-incremental-moment-matching}
  {Overcoming catastrophic forgetting by incremental moment matching}.
\newblock In \emph{NIPS}, pages 4652--4662.

\bibitem[{Liu(2015)}]{liu2015sentiment}
Bing Liu. 2015.
\newblock \emph{Sentiment analysis: Mining opinions, sentiments, and emotions}.
\newblock Cambridge University Press.

\bibitem[{Liu et~al.(2015)Liu, Gao, Liu, and Zhang}]{liu2015automated}
Qian Liu, Zhiqiang Gao, Bing Liu, and Yuanlin Zhang. 2015.
\newblock Automated rule selection for aspect extraction in opinion mining.
\newblock In \emph{IJCAI}.

\bibitem[{Lopez{-}Paz and Ranzato(2017)}]{Lopez2017gradient}
David Lopez{-}Paz and Marc'Aurelio Ranzato. 2017.
\newblock \href
  {http://papers.nips.cc/paper/7225-gradient-episodic-memory-for-continual-learning}
  {Gradient episodic memory for continual learning}.
\newblock In \emph{NIPS}, pages 6467--6476.

\bibitem[{Lv et~al.(2019)Lv, Wang, Liu, Chen, and
  Zhang}]{DBLP:conf/dasfaa/LvWLCZ19}
Guangyi Lv, Shuai Wang, Bing Liu, Enhong Chen, and Kun Zhang. 2019.
\newblock \href {https://doi.org/10.1007/978-3-030-18576-3\_47} {Sentiment
  classification by leveraging the shared knowledge from a sequence of
  domains}.
\newblock In \emph{DASFAA}, pages 795--811.

\bibitem[{Mallya and Lazebnik(2018)}]{Mallya2017packnet}
Arun Mallya and Svetlana Lazebnik. 2018.
\newblock \href {https://doi.org/10.1109/CVPR.2018.00810} {Packnet: Adding
  multiple tasks to a single network by iterative pruning}.
\newblock In \emph{CVPR}, pages 7765--7773.

\bibitem[{McCloskey and Cohen(1989)}]{mccloskey1989catastrophic}
Michael McCloskey and Neal~J Cohen. 1989.
\newblock Catastrophic interference in connectionist networks: The sequential
  learning problem.
\newblock In \emph{Psychology of learning and motivation}, volume~24, pages
  109--165. Elsevier.

\bibitem[{Merchant et~al.(2020)Merchant, Rahimtoroghi, Pavlick, and
  Tenney}]{DBLP:journals/corr/abs-2004-14448}
Amil Merchant, Elahe Rahimtoroghi, Ellie Pavlick, and Ian Tenney. 2020.
\newblock \href {http://arxiv.org/abs/2004.14448} {What happens to {BERT}
  embeddings during fine-tuning?}
\newblock \emph{CoRR}, abs/2004.14448.

\bibitem[{Parisi et~al.(2019)Parisi, Kemker, Part, Kanan, and
  Wermter}]{Parisi2019continual}
German~Ignacio Parisi, Ronald Kemker, Jose~L. Part, Christopher Kanan, and
  Stefan Wermter. 2019.
\newblock \href {https://doi.org/10.1016/j.neunet.2019.01.012} {Continual
  lifelong learning with neural networks: {A} review}.
\newblock \emph{Neural Networks}, 113:54--71.

\bibitem[{Qin et~al.(2020)Qin, Hu, and Liu}]{qin2020using}
Qi~Qin, Wenpeng Hu, and Bing Liu. 2020.
\newblock Using the past knowledge to improve sentiment classification.
\newblock In \emph{Proceedings of the 2020 Conference on Empirical Methods in
  Natural Language Processing: Findings}, pages 1124--1133.

\bibitem[{Rebuffi et~al.(2017)Rebuffi, Kolesnikov, Sperl, and
  Lampert}]{Rebuffi2017}
Sylvestre{-}Alvise Rebuffi, Alexander Kolesnikov, Georg Sperl, and Christoph~H.
  Lampert. 2017.
\newblock \href {https://doi.org/10.1109/CVPR.2017.587} {icarl: Incremental
  classifier and representation learning}.
\newblock In \emph{CVPR}, pages 5533--5542.

\bibitem[{Rostami et~al.(2019)Rostami, Kolouri, and Pilly}]{Rostami2019ijcai}
Mohammad Rostami, Soheil Kolouri, and Praveen~K. Pilly. 2019.
\newblock \href {https://doi.org/10.24963/ijcai.2019/463} {Complementary
  learning for overcoming catastrophic forgetting using experience replay}.
\newblock In \emph{IJCAI}, pages 3339--3345.

\bibitem[{Ruvolo and Eaton(2013)}]{ruvolo2013ella}
Paul Ruvolo and Eric Eaton. 2013.
\newblock \href {http://proceedings.mlr.press/v28/ruvolo13.html} {{ELLA:} an
  efficient lifelong learning algorithm}.
\newblock In \emph{ICML}, pages 507--515.

\bibitem[{Sabour et~al.(2017)Sabour, Frosst, and Hinton}]{sabour2017dynamic}
Sara Sabour, Nicholas Frosst, and Geoffrey~E Hinton. 2017.
\newblock Dynamic routing between capsules.
\newblock In \emph{NIPS}, pages 3856--3866.

\bibitem[{Seff et~al.(2017)Seff, Beatson, Suo, and Liu}]{Seff2017continual}
Ari Seff, Alex Beatson, Daniel Suo, and Han Liu. 2017.
\newblock \href {http://arxiv.org/abs/1705.08395} {Continual learning in
  generative adversarial nets}.
\newblock \emph{CoRR}, abs/1705.08395.

\bibitem[{Serr{\`{a}} et~al.(2018)Serr{\`{a}}, Suris, Miron, and
  Karatzoglou}]{Serra2018overcoming}
Joan Serr{\`{a}}, Didac Suris, Marius Miron, and Alexandros Karatzoglou. 2018.
\newblock \href {http://proceedings.mlr.press/v80/serra18a.html} {Overcoming
  catastrophic forgetting with hard attention to the task}.
\newblock In \emph{ICML}, pages 4555--4564.

\bibitem[{Shin et~al.(2017)Shin, Lee, Kim, and Kim}]{Shin2017continual}
Hanul Shin, Jung~Kwon Lee, Jaehong Kim, and Jiwon Kim. 2017.
\newblock \href
  {http://papers.nips.cc/paper/6892-continual-learning-with-deep-generative-replay}
  {Continual learning with deep generative replay}.
\newblock In \emph{NIPS}, pages 2990--2999.

\bibitem[{Shu et~al.(2017)Shu, Xu, and Liu}]{ShuXuLiu2017}
Lei Shu, Hu~Xu, and Bing Liu. 2017.
\newblock \href {https://doi.org/10.18653/v1/P17-2023} {Lifelong learning {CRF}
  for supervised aspect extraction}.
\newblock In \emph{ACL}, pages 148--154.

\bibitem[{Silver et~al.(2013)Silver, Yang, and Li}]{Silver2013}
Daniel~L. Silver, Qiang Yang, and Lianghao Li. 2013.
\newblock \href {http://www.aaai.org/ocs/index.php/SSS/SSS13/paper/view/5802}
  {Lifelong machine learning systems: Beyond learning algorithms}.
\newblock In \emph{Lifelong Machine Learning, Papers from the 2013 {AAAI}
  Spring Symposium, Palo Alto, California, USA, March 25-27, 2013}.

\bibitem[{Sun et~al.(2019)Sun, Huang, and Qiu}]{sun-etal-2019-utilizing}
Chi Sun, Luyao Huang, and Xipeng Qiu. 2019.
\newblock \href {https://doi.org/10.18653/v1/N19-1035} {Utilizing {BERT} for
  aspect-based sentiment analysis via constructing auxiliary sentence}.
\newblock In \emph{NAACL}, pages 380--385, Minneapolis, Minnesota. Association
  for Computational Linguistics.

\bibitem[{Tang et~al.(2016)Tang, Qin, and Liu}]{tang-etal-2016-aspect}
Duyu Tang, Bing Qin, and Ting Liu. 2016.
\newblock \href {https://doi.org/10.18653/v1/D16-1021} {Aspect level sentiment
  classification with deep memory network}.
\newblock In \emph{EMNLP}, pages 214--224, Austin, Texas. Association for
  Computational Linguistics.

\bibitem[{Vaswani et~al.(2017)Vaswani, Shazeer, Parmar, Uszkoreit, Jones,
  Gomez, Kaiser, and Polosukhin}]{vaswani2017attention}
Ashish Vaswani, Noam Shazeer, Niki Parmar, Jakob Uszkoreit, Llion Jones,
  Aidan~N Gomez, {\L}ukasz Kaiser, and Illia Polosukhin. 2017.
\newblock Attention is all you need.
\newblock In \emph{NeurIPS}, pages 5998--6008.

\bibitem[{Wang et~al.(2019)Wang, Liu, Wang, Ma, and Yang}]{hao2019forward}
Hao Wang, Bing Liu, Shuai Wang, Nianzu Ma, and Yan Yang. 2019.
\newblock \href {http://proceedings.mlr.press/v101/wang19f.html} {Forward and
  backward knowledge transfer for sentiment classification}.
\newblock In \emph{ACML}, pages 457--472.

\bibitem[{Wang et~al.(2018)Wang, Lv, Mazumder, Fei, and
  Liu}]{shuai2018lifelong}
Shuai Wang, Guangyi Lv, Sahisnu Mazumder, Geli Fei, and Bing Liu. 2018.
\newblock \href {https://doi.org/10.1109/BigData.2018.8622304} {Lifelong
  learning memory networks for aspect sentiment classification}.
\newblock In \emph{{IEEE} International Conference on Big Data}, pages
  861--870.

\bibitem[{Xia et~al.(2017)Xia, Jiang, and He}]{xia2017distantly}
Rui Xia, Jie Jiang, and Huihui He. 2017.
\newblock \href {https://doi.org/10.1109/TAFFC.2017.2771234} {Distantly
  supervised lifelong learning for large-scale social media sentiment
  analysis}.
\newblock \emph{{IEEE} Trans. Affective Computing}, 8(4):480--491.

\bibitem[{Xu et~al.(2019)Xu, Liu, Shu, and Yu}]{DBLP:conf/naacl/XuLSY19}
Hu~Xu, Bing Liu, Lei Shu, and Philip~S. Yu. 2019.
\newblock \href {https://doi.org/10.18653/v1/n19-1242} {{BERT} post-training
  for review reading comprehension and aspect-based sentiment analysis}.
\newblock In \emph{NAACL-HLT}, pages 2324--2335. Association for Computational
  Linguistics.

\bibitem[{Xu et~al.(2018)Xu, Xie, Shu, and Yu}]{Xu2018pro}
Hu~Xu, Sihong Xie, Lei Shu, and Philip~S. Yu. 2018.
\newblock Dual attention network for product compatibility and function
  satisfiability analysis.
\newblock In \emph{AAAI}.

\bibitem[{Xue and Li(2018)}]{DBLP:conf/acl/LiX18}
Wei Xue and Tao Li. 2018.
\newblock \href {https://doi.org/10.18653/v1/P18-1234} {Aspect based sentiment
  analysis with gated convolutional networks}.
\newblock In \emph{ACL}, pages 2514--2523. Association for Computational
  Linguistics.

\bibitem[{Zeng et~al.(2019)Zeng, Chen, Cui, and Yu}]{zeng2019continuous}
Guanxiong Zeng, Yang Chen, Bo~Cui, and Shan Yu. 2019.
\newblock Continuous learning of context-dependent processing in neural
  networks.
\newblock \emph{Nature Machine Intelligence}.

\bibitem[{Zhao et~al.(2019)Zhao, Peng, Eger, Cambria, and
  Yang}]{DBLP:conf/acl/ZhaoPECY19}
Wei Zhao, Haiyun Peng, Steffen Eger, Erik Cambria, and Min Yang. 2019.
\newblock \href {https://doi.org/10.18653/v1/p19-1150} {Towards scalable and
  reliable capsule networks for challenging {NLP} applications}.
\newblock In \emph{ACL}, pages 1549--1559. Association for Computational
  Linguistics.

\end{thebibliography}
\bibliographystyle{acl_natbib}

\end{document}